# Vehicle Acceleration Prediction Considering Environmental Influence and Individual Driving Behavior

Wenxuan Wang[a], Lexing Zhang[a,b,*], Jiale Lei[a], Yin Feng[a], Hengxu Hu[a]

[a]Chang'an University, Xi'an, Shaanxi, China; [b]National University of Singapore, Singapore, Singapore.

**ABSTRACT**

Accurate vehicle acceleration prediction is critical for intelligent driving control and energy efficiency management, particularly in environments with complex driving behavior dynamics. This paper proposes a general short-term vehicle acceleration prediction framework that jointly models environmental influence and individual driving behavior. The framework adopts a dual-input design by incorporating environmental sequences, constructed from historical traffic variables such as percentile-based speed and acceleration statistics of multiple vehicles at specific spatial locations, capture group-level driving behavior influenced by the traffic environment. In parallel, individual driving behavior sequences represent motion characteristics of the target vehicle prior to the prediction point, reflecting personalized driving styles. These two inputs are processed using an LSTM Seq2Seq model enhanced with an attention mechanism, enabling accurate multi-step acceleration prediction. To demonstrate the effectiveness of the proposed method, an empirical study was conducted using high-resolution radar–video fused trajectory data collected from the exit section of the Guangzhou Baishi Tunnel. Drivers were clustered into three categories—conservative, moderate, and aggressive—based on key behavioral indicators, and a dedicated prediction model was trained for each group to account for driver heterogeneity.Experimental results show that the proposed method consistently outperforms four baseline models (RNN, ANN, CNN, BiLSTM), yielding a 10.9% improvement in accuracy with the inclusion of historical traffic variables and a 33% improvement with driver classification. Although prediction errors increase with forecast distance, incorporating environment- and behavior-aware features significantly enhances model robustness. The findings provide new insights into vehicle acceleration prediction and offer valuable implications for autonomous vehicle control and energy management systems.

*Keywords: Traffic Safety, Acceleration Prediction, Seq2Seq, Driving behavior, Environmental influence*

# 1. Introduction

Driving behavior prediction plays a crucial role in modern intelligent transportation systems (ITS), such as advanced driver assistance systems(ADAS) and other safety-related applications(Shi and Abdel-Aty 2015). By forecasting driving behavior, these systems can assess a vehicle's real-time risk level and deploy early alert systems and targeted interventions. Furthermore, driving behavior prediction contributes to optimizing vehicle energy management, for example, powertrain control strategies for plug-in hybrid electric vehicles(PHEVs) and other electrified vehicles rely on accurate prediction of future driving behaviors. Effective vehicle behavior prediction can thus significantly enhance overall powertrain efficiency and reduce fuel consumption(Jiang and Fei 2014).

*Corresponding author.
 Email address: 2021902169@chd.edu.cn (Lexing Zhang)

Despite these benefits, accurately predicting the behavior of an individual vehicle remains highly challenging. Vehicles operate within a complex human-vehicle-road-environment system, where the behavior is influenced by myriad factors, including prevailing traffic conditions, vehicle types, road geometry, and driver characteristics. In particular, driver heterogeneity leads to substantial variability in driving behavior under complex traffic scenarios, complicating predictive efforts.

Existing nonparametric models—such as neural networks (NNs) and nonparametric regression—have demonstrated effectiveness in predicting aggregate traffic behavior using data collected from roadside loop detectors and speed sensors(Ma et al. 2015). However, these sensors only provide macroscopic traffic flow information rather than individual vehicle trajectories. As a result, such models perform poorly in predicting individual vehicle speed profiles, which require microscopic behavioral resolution. Recent advances in roadside sensing technologies—including millimeter-wave radar and video-based tracking—now enable continuous, long-range collection of individual vehicle trajectories. To fully leverage these rich trajectory datasets for precise driving behavior prediction, novel data-driven algorithms must be developed and implemented.

This paper proposes a novel vehicle acceleration prediction model based on an attention-enhanced long short-term memory (LSTM) sequence-to-sequence (Seq2Seq) network. The process begins by using high-precision vehicle trajectory data to explore the acceleration and deceleration behaviors caused by sudden environmental changes. Next, considering individual driver differences, drivers are classified into distinct groups based on their driving styles. Then, acceleration models are built for each group. These models use statistical features extracted from historical traffic data to represent the influence of the environment, combined with individual vehicle trajectory information reflecting the driver's style, to predict the vehicle's acceleration. Finally, the effectiveness of the model is validated. The main contributions of this study are summarized:

1) **Driving behavior analysis**: Using high-precision vehicle trajectory data, the study explores the impact of dramatic environmental changes on vehicle behaviors. The findings provide both empirical data and a theoretical foundation for accurately predicting driver behavior, improving traffic safety and behavior modeling.

2) **Innovative dual-input model framework:** The proposed model introduces a novel input structure that simultaneously incorporates environmental influence and individual driving behavior. Environmental influences are represented through environmental sequences constructed from historical traffic variables of multiple vehicles, while individual driving sequences capture the unique motion patterns of the target vehicle. This dual-input approach significantly enhances prediction accuracy by addressing both dynamic traffic conditions and driver heterogeneity.

3) **Attention-enhanced LSTM Seq2Seq framework:** The attention mechanism in this model allows it to focus more on segments of the sequence exhibiting greater fluctuations, particularly critical locations. By integrating the Seq2Seq framework, the model simultaneously predicts acceleration at multiple future positions, generating a multi-step prediction sequence in a single pass. This reduces error accumulation, enhances computational efficiency and achieves superior predictive performance. As a result, the proposed approach outperforms baseline models including Recurrent Neural Network(RNN), Artificial Neural Network(ANN), Convolutional Neural Net-works(CNN), and conventional Long Short-Term Memory(LSTM) models.

The organization of the paper is as follows. Section 2 reviews the literature; Section 3 introduces data sources. Section 4 outlines the methods employed in this study, forming the basis for the short-term acceleration prediction model; and Section 5 presents the performance evaluation of the prediction model, Section 6 discusses the key findings and their implications. Finally, Section 7 concludes the paper and outlines directions for future research.

## 2. Literature review

Vehicle trajectory prediction encompasses both speed and acceleration prediction, which are widely used in motion planning, autonomous vehicle control, risk warning and power management for electric vehicles. This section reviews the relevant literature on driving behavior prediction (Section 2.1) methods and then examine common data sources(Section 2.2), before identifying the specific gap addressed by this study.

## 2.1 Driving behavior prediction

A wide range of data-driven techniques has been applied to traffic parameter prediction, from classical approaches (Kalman filter(Williams and Hoel 2003), Support Vector Machines (SVM)(Sun, Zhang, and Zhang 2017), Random Forest(RF)(Ma et al. 2015), Hidden Markov Models (HMM) (Jiang and Fei 2017) to modern deep learning architectures, such as Long Short-Term Memory (LSTM)(Ma et al. 2015), Convolutional Neural Networks (CNN)(Lv et al. 2014), Transformer(Yu, Shi, and Zhang 2023), Deep Reinforcement Learning (DRL)(Wei et al. 2018) and Graph Neural Networks (GNN)(Li et al. 2017). These methods improve prediction accuracy by capturing complex interactions and long-term dependencies. However, conventional RNN/LSTM models rely on recursive one-step-ahead forecasting, which amplifies errors over multiple steps and cannot process long sequences in parallel(Bengio et al. 2015; Vaswani 2017). Sequence-to-Sequence (Seq2Seq) architectures address these shortcomings by generating multi-step outputs in a single pass, thereby mitigating the issue of error accumulation in recursive methods. The introduction of the attention mechanism enables the Seq2Seq model to dynamically focus on the most informative time segments, significantly reducing error propagation in long-horizon predictions (Luong 2015).

In addition to the continuous development of prediction models, some studies have focused on improving prediction performance by partitioning data into contextual domains before prediction modeling. For instance, different algorithms can be optimized for specific domains: NIGA-SVM is suitable for urban areas, while GA-SVM is effective for freeways and suburban regions(Li et al. 2018). Furthermore, clustering acceleration data and other parameters that capture interactions among traffic can enhance prediction accuracy by applying different models and parameters to each cluster(Luo, Hu, and Huang 2023; Zou et al. 2022; Tang et al. 2024). This indicates that the combination of clustering and deep learning techniques can improve acceleration prediction accuracy(Luo, Hu, and Huang 2023). Therefore, this paper explores the clustering of driving styles based on trajectories prior to applying acceleration prediction.

While previous research has sought to achieve more accurate predictions by refining models, it is important to recognize that vehicle operation is influenced by a complex system comprising human, vehicle, road, and environmental factors. Effectively extracting parameters that represent these influences is key to improving prediction performance. Consequently, researchers have incorporated various feature variables as additional inputs to enhance vehicle speed prediction accuracy. For example, Yeon et al. (Yeon et al. 2019) included internal vehicle information, relative speed and distance to the vehicle ahead, and the ego-vehicle's location. Xu et al. (Xu et al. 2022) utilized driver pupil information and illumination levels in tunnels. Yuan et al.(Yuan et al. 2023) considered historical speed, traffic participant information, and vehicle position. Jones and Han (Jones and Han 2019) accounted for the acceleration and speed of the ego vehicle, as well as the relative speed and distance to surrounding vehicles. Su et al. (Su et al. 2020) included vehicle lane, class, speed, acceleration, and relative distance from three nearest front neighbors. In this paper, statistical features of multiple passing vehicles will be integrated into the prediction model to represent the impact of road and environmental conditions on vehicle behavior.

In summary, this paper proposes a Seq2Seq-based vehicle acceleration prediction method. Driving styles are classified and separate acceleration prediction models are developed for each driver group. Within each model, environmental influences on traffic and individual driving behavior are both considered.

## 2.2 Data type and acquisition methods

Most existing driving behavior prediction relies on aggregated traffic flow data collected by loop detectors. Recent advances in sensing technologies, however, now enable direct collection of individual trajectories. Test-vehicle experiments(Yuan et al. 2023; Yeon et al. 2019) provide high-precision motion data of ego vehicle motion and surrounding vehicle information (Yeon et al. 2019). Additionally, roadside surveillance systems capture trajectory along specific road segments(Jiang and Fei 2017), for example, NGSIM naturalistic dataset is limited to basic freeway sections that offers trajectory of individual vehicle and surrounding vehicles. Occasionally, simulation data are utilized to supplement real-world data(Jiang and Fei 2017).

Recently, radar sensing enables the continuous tracking of vehicle trajectories along extensive road segments, overcoming challenges posed by low illumination and adverse weather conditions. When fused with video data, radar's shortcomings in vehicle classification and distance estimation are mitigated, producing high-fidelity trajectory datasets that span entire expressway sections (Wang et al. 2024; Yan, Giorgetti, and Paolini 2021; Yan et al. 2022; Yan et al. 2021). Therefore, this paper leverages large-scale, radar–camera fused trajectory data collected in Guangzhou expressway tunnels to develop and validate a tunnel-specific vehicle acceleration prediction model.

# 3. Data processing

This section describes the data collection and processing procedures for vehicle acceleration prediction, based on millimeter-wave radar and video surveillance in the Baishi Tunnel on the Guangzhou-Lianyang Expressway, China.

## 3.1 Study site and data-collection period

The Baishi Tunnel is located on the Guangzhou-Lianyang Expressway in Guangdong, China. It is a two-way, six-lane tunnel with a clear width of 14 meters and a clear height of 5 meters. It has a total length of 1,800 meters with the posted speed limit of 100 km/h. The uniform geometry and constrained environment make it ideal for examining vehicle responses to changing tunnel conditions, particularly at the tunnel exit. Data were collected over 700m segment spanning 380m inside to 320m outside the tunnel exit portal. Data collection took place on September 30, 2023, the first day of China's National Day Holiday. A total of 400 minutes of data were recorded from 11:35 a.m. to 6:15 p.m., continuously tracking 3,847 vehicles.

## 3.2 Equipment and measurement settings

Fourteen millimeter-wave radars were mounted along the Baishi Tunnel (See **Figure1(a)**) to capture vehicle trajectories. Each radar has a coverage range of up to 400 meters and is spaced approximately 130 meters apart. The radar detection frequency is set at 20 Hz, allowing for continuous, real-time tracking of vehicle position, speed, and acceleration. In addition, two high-definition video cameras were positioned near the tunnel exit and entrance to capture vehicle physical characteristics, such as type, color, as well as movement behavior. The cameras recorded in 1080p at 60 frames per second (fps)(See **Figure2** ), with the frame rate reduced to 20 fps during post-processing to facilitate synchronization with the radar data.

## 3.3 Trajectory extraction and fusion

Individual vehicle trajectories were extracted using a Track-Before-Detect (TBD) algorithm(Yan et al. 2019), which ensured stable and continuous tracking throughout the tunnel. The resulting radar-derived vehicle positions at various timestamps are presented in **Figure1**. To further improve data quality, vehicle attributes—including type, color, and lane position—were manually verified against video footage. This step helped correct common radar errors such as missed detections, incomplete speed profiles, or trajectory merging of adjacent vehicles(**Figure1(b)**). Although the volume of video data was relatively small, it was manually verified in this study. Future work will automate this step using YOLO-based object detection(Raza et al. 2024)and radar–vision fusion techniques to improve accuracy and scalability.

## 3.4 Reliability of the dataset

Compared to video-only vehicle trajectory datasets (e.g. the NGSIM("NGSIM [WWW Document]" 2006) and HighD(Krajewski 2018) datasets), the radar–camera fusion approach used in this study offers several significant advantages, particularly in complex and enclosed environments such as expressway tunnels. (1) Enhanced measurement precision: Radar accurate captures distance, speed, and acceleration, while video provides semantic attributes such as vehicle type and color. Fusion mitigates occlusion and lighting limitations of video-only systems. (2) Robustness in dynamic traffic: Radar performs reliably in multi-lane and congested traffic. Fusion further improves tracking by cross-validating detections with visual cues. (3) Environmental resilience: Radar maintains stability in low-light and enclosed environments, such as tunnels. Video adds semantic context, complementing radar's spatial measurements. (4) Real-time high-frequency sampling: Radar offers high-frequency (20 Hz) real-time tracking, and synchronization with video enhances both spatial detail and temporal consistency.

Although radar alone cannot capture fine-grained visual features or accurate lane positions, video integration effectively compensates for these limitations. As a result, the fused system provides high-fidelity, continuous vehicle trajectories well suited for acceleration prediction in challenging road environments.

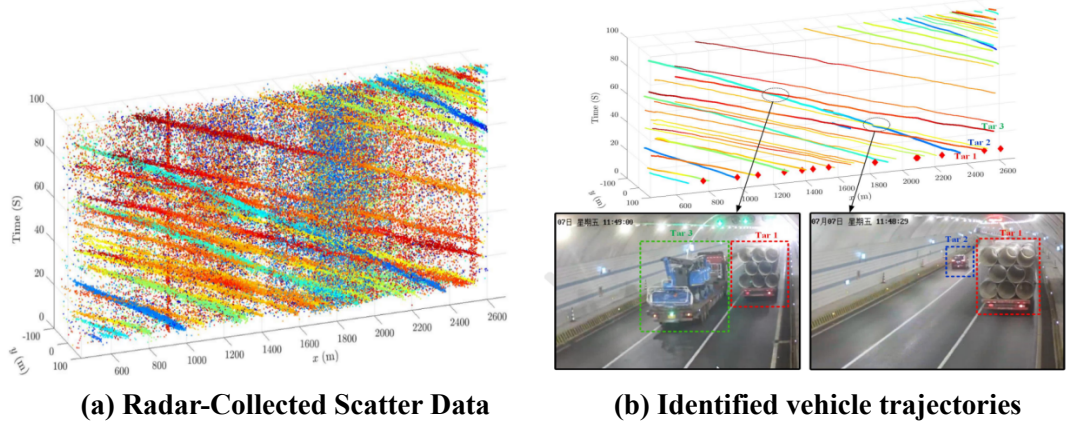

**(a) Radar-Collected Scatter Data**   **(b) Identified vehicle trajectories**

**Figure 1 Data acquisition**

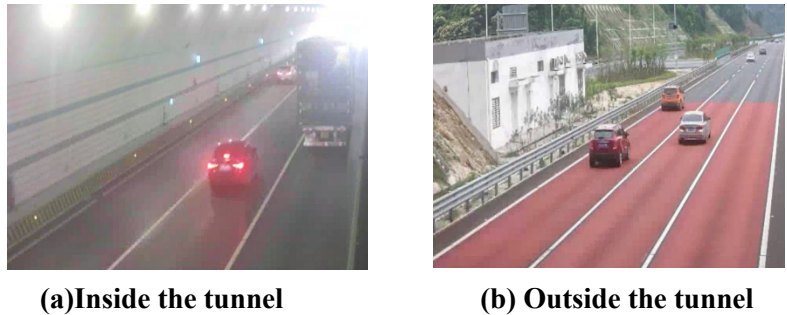

**(a) Inside the tunnel**   **(b) Outside the tunnel**

**Figure 2 Traffic flow near the tunnel exit captured by the camera**

## 3.5 Data processing

To obtain the input required for the prediction model, the speed-time data is numerically differentiated to obtain acceleration-time data. Due to the high sampling frequency, the numerical noise and temporal lag introduced by differentiation are minimal and have negligible impact on prediction performance.

In this paper, the vehicle's acceleration is assumed to vary linearly within each sampling interval of 0.05 seconds (corresponding to 20Hz). Based on this assumption, a linear interpolation algorithm is used to estimate the vehicle's acceleration at specific spatial locations along the trajectory. Considering the average distance a vehicle travels within each 0.05-second sampling interval, the interpolation interval is set to 1 meter to ensure spatial consistency in trajectory processing.

The acceleration at specific position $a_{LI}$ can be calculated by Equation (1).

$$a_{LI} = a_t + \frac{a_t - a_{t-1}}{x_t - x_{t-1}}(x_{LI} - x_t) \tag{1}$$

where $a$ represents the vehicle acceleration, $x$ denotes the vehicle position, $t$ and $t-1$ indicate the nearest existing data points before and after the position to be interpolated, The term $LI$ (Linear Interpolation) is used to refer to the position to be interpolated.

## 4. Prediction Framework and Model Design

This section provides an overview of the short-term acceleration prediction framework. The process of classifying driving characteristics is then described, followed by the introduction of the historical traffic variables used in the model. Finally, a detailed explanation of the deep learning acceleration prediction model.

## 4.1 Overview of vehicle acceleration prediction framework

This study proposes a three-stage short-term acceleration prediction framework. The framework consists of data preprocessing, model construction, and result evaluation. The overall workflow is illustrated in **Figure 3**.

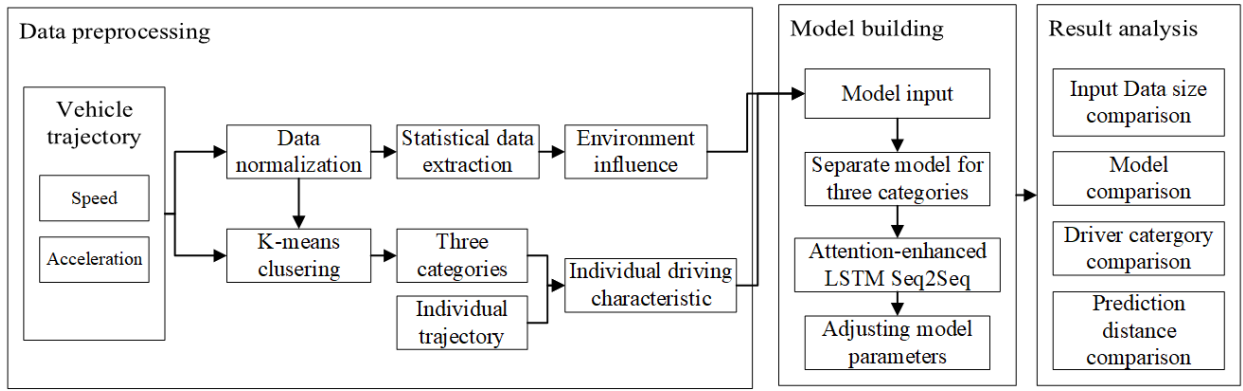

**Figure 3 Flow chart of the vehicle acceleration prediction**

**Module 1: Data Preprocessing.** Vehicle trajectory data including speed and acceleration is extracted and denoised. The cleaned data is then used to analyze the influence of the environment on vehicle behavior, based on traffic historical variables. These variables are statistically summarized to construct the first model input, referred to as environmental sequence. To capture individual driving behavior, vehicle-level features, such as average speed, average acceleration, and acceleration range are extracted and input to K-means clustering algorithm for driver classification. The resulting driving categories and their corresponding trajectories form the second model input, termed the individual sequence.

**Module 2: Model Building.** Separate acceleration prediction models are developed for each driver category, using the dual-input framework in Module 1. The models are constructed based on an attention-enhanced LSTM Seq2Seq framework, which enables the network to focus on key temporal and environmental features during prediction. Model parameters are optimized during training by minimizing prediction errors. The dataset is split into 70% for training, 20% for testing, and 10% for validation.

**Module 3: Result Analysis.** The prediction results are evaluated across three key dimensions. First, the proposed model is compared against four baseline models (BiLSTM, RNN, ANN, and CNN) to assess its accuracy. Second, the impact of different data size on model performance is analyzed to determine an appropriate input size for capturing the historical traffic variables. Finally, model performance is evaluated across different driver categories to investigate the effect of driving style variability on predictive performance.

## 4.2 Environmental and individual input design

Vehicle behavior is shaped by both environmental influences and individual driver variability. For example, on curved or constrained roads, most vehicles tend to decelerate, yet the degree of deceleration differs significantly across drivers. This reflects a common environmental effect paired with individual-specific responses. The environmental influence can be understood as a relatively deterministic trend shared by multiple vehicles under similar conditions, whereas individual driving styles introduce stochastic variations around this trend. In this study, these two aspects are modeled separately.

(1) Environmental sequence

The deterministic impact of the environment is represented through historical traffic variables, specifically the 20th, 40th, 60th, and 80th percentile values of speed and acceleration at specific tunnel locations. These percentile-based features capture collective driving behavior shaped by the spatial environment, and are used as part of the input to the acceleration prediction model.

(2) Individual driving sequence

In contrast, the individual driving behavior sequence reflects the stochastic variability introduced by the unique driving style of the target vehicle. This sequence includes the historical speed and acceleration values of the vehicle prior to the prediction point, capturing recent motion patterns. To further characterize driver-specific behavior, unsupervised k-means clustering based on acceleration range, average speed, and average acceleration for each vehicle is used to classify drivers into three categories: conservative, moderate, and aggressive. This classification is used to build dedicated prediction models for each group, enabling the model to better account for driver heterogeneity.

The objective of the k-means algorithm is to minimize the sum of the squared distances between the sample points within a cluster and the cluster center. The data set presented in this paper is designated as $X = \{x_1, x_2, \ldots, x_n\}$, where $x_i$ represents a data object. This objective function can be expressed in the Equation (2).

$$min \, W(C) = \sum_{i=1}^{k} \sum_{x \in C_i} \| x_j - \mu_i \|^2 \tag{2}$$

where $k$ is the total number of clusters; $C_i$ is the $i$th cluster, $x_j$ is the $j$th data point in cluster $C_i$; $\mu_i$ is the center of cluster $C_i$.

While the k-means algorithm is highly effective for cluster analysis, it is sensitive to the choice of the number of clusters k. Therefore, selecting an appropriate number of clusters is cruicial for improving prediction accuracy. In this paper, the model selection tools such as the Akaike Information Criterion (AIC) and Bayesian Information Criterion (BIC) are used to determine the optimal number of clusters. Both AIC and BIC are based on the principle of penalized likelihood, penalizing models with more parameters; however, BIC imposes a heavier penalty than AIC. As a result, BIC tends to favor simpler models, whereas AIC is inclined towards more complex models. The calculation for AIC and BIC can be expressed in Equation (3)-(4).

$$AIC = 2k - 2\ln(L) \tag{3}$$

$$BIC = k * \ln(n) - 2\ln(L) \tag{4}$$

where $n$ is the sample size, and $L$ is the likelihood function of the k-means algorithm.

To calculate the Euclidean distance between a given point and the centroid of a specific cluster, the following Equation (5) can be used.

$$D(x_i, x_j) = \sqrt{\sum_{r=1}^{d} (x_i - x_j)^2} \, (i = 1,2,\dots,n; j = 1,2,\dots,k) \tag{5}$$

where $x_i$ represents a point within the class; $x_j$ represents category centers; $n$ represents the total number of data; and $r$ represents the data dimension.

As demonstrated by Equation (5), the value of D is represented by the $n \times k$ matrix. In this paper, we employ the mean deviation as a representation of the likelihood function for k-means clustering, as detailed in Equation (6).

$$L = \frac{1}{n} \left[ \sum_{i=1}^{n} \left( \min_{1 \le i \le n} D_{n \times k} \right) \right] \tag{6}$$

where $i$ represents a row of the data and $L$ is the likelihood function.

In order to eliminate the effect of magnitude in when calculating AIC and BIC for different values of k, L is normalized and $\ln(L')$ is normalized, as demonstrated in Equation (7).

$$\ln(L') = \ln \left[ \frac{L - L_{min}}{L_{max} - L_{min}} (e - 1) + 1 \right] \tag{7}$$

In order to calculate the two terms before and after on an equivalent scale, it is necessary to multiply $2\ln(L')$ by 10, resulting in Equations (8) and (9).

$$AIC = 2 \times k + 2 \times 10 \times ln(L') \qquad (8)$$

$$BIC = k * ln(n) \times 10 \times ln(L') \qquad (9)$$

## 4.3 Short-term acceleration prediction model

This section introduces the acceleration prediction model based on the attention-enhanced LSTM Seq2Seq framework. By combining the attention mechanism with the Seq2Seq structure, the model can simultaneously predict the acceleration at multiple future positions in a single step, thereby reducing error accumulation and improving computational efficiency. The attention mechanism enables the model to dynamically assign weights to different parts of the input sequence, allowing it to focus more on locations that exhibit significant behavioral variations.

The Seq2Seq framework, originally developed for tasks such as machine translation and text summarization, consists of two core components: an Encoder and a Decoder. The Encoder processes the input sequence through stacked recurrent layers (typically LSTM units), generating a context vector that summarizes the input. The Decoder then transforms this context vector into an output sequence using a second set of recurrent layers.

However, traditional Seq2Seq models often face difficulties when handling long input sequences, as compressing all input information into a single context vector may lead to information loss and degraded performance. To overcome this limitation, an attention mechanism is introduced. Instead of relying solely on the fixed-length context vector, the attention-enhanced Decoder selectively focuses on different parts of the input sequence during each decoding step. This enables the model to capture more detailed temporal dependencies and improves prediction accuracy for complex driving behaviors. The computational process is detailed in Equations (10) - (12).

$$E_t = \tanh(\text{attn}(s_{t-1}, H)) \qquad (10)$$

$$\widetilde{a_t} = vE_t \qquad (11)$$

$$a_t = \text{softmax}(\widetilde{a_t}) \qquad (12)$$

where s is the initial hidden state derived from the encoder output via a fully connected layer; H is the output obtained by the encoder after encoding the input sequence; attn() is a fully connected neural network.

In **Figure5,** the environmental sequence and individual vehicle sequence serve as the inputs to the prediction model. However, due to differences in sequence lengths relative to the prediction target, these inputs cannot be directly fed into the Seq2Seq network. To address this mismatch, a fully connected neural network layer is applied to the environmental sequence to align its dimensionality with that of the individual driving sequence, allowing both to be jointly processed by the encoder. The resulting multivariate inputs are then fed into the encoder of the Seq2Seq model. The encoder's final hidden state is passed through the attention layer, which generates a context vector used to initialize the decoder's hidden state ($h0$). The decoder subsequently produces the multi-step acceleration prediction output.

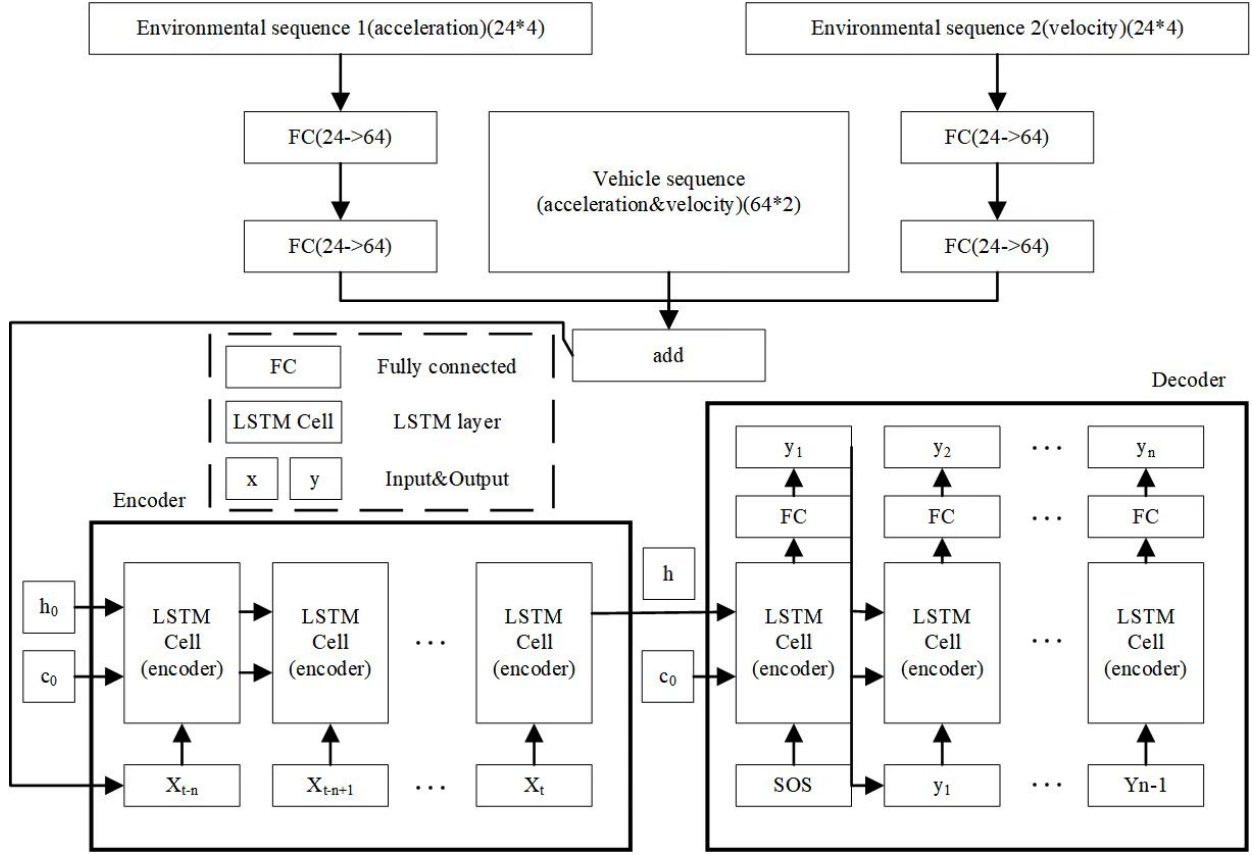

**Figure 5** Architecture of the Attention-Enhanced LSTM Seq2Seq Network

The length of the historical sequence significantly affects the performance of prediction models. Short time series may lead to incomplete contextual information, thereby reducing prediction accuracy. Conversely, an excessively long sequences increase model complexity and computational cost, which may hinder practical applicability. To balance prediction accuracy and efficiency, this study adopts a historical input length of 100 meters.

The data is randomly divided into 70% for training, 20% for testing, and 10% for validation. A dropout mechanism is also employed during training to improve generalization by randomly deactivating a subset of neurons, thereby reducing the risk of overfitting.

To accurately assess the model's performance, comparisons are made against several baseline models. Mean Absolute Error (MAE) and Root Mean Squared Error (RMSE) are used as evaluate metrics. MAE measures the average magnitude of errors between predicted and actual values, while RMSE represents the square root of the average squared errors. These metrics are defined as follows:

$$\text{MAE} = \frac{1}{m} \sum_{i=1}^{m} |y_i - \hat{y}_i| \tag{13}$$

$$\text{RMSE} = \sqrt{\frac{1}{m}\sum_{i=1}^{m}(y_i - \hat{y}_i)^2} \tag{14}$$

where y is the true value of the variable to be predicted and $\hat{y}$ is the predicted value of the variable to be predicted. m represents the total predicted length and i represents a certain predicted location.

## 5. Results

In this section, the distribution characteristics of vehicle speed and acceleration at the tunnel exit are first analyzed. Then, the results of driver clustering based on speed and acceleration information are presented. Subsequently, the performance of the vehicle acceleration prediction model is evaluated from multiple aspects.

## 5.1 Vehicle acceleration and speed characteristics

The distribution of vehicle acceleration near the tunnel exit was obtained from the collected trajectory data of all vehicles, as shown in **Figure 6.** The y-axis represents the vehicle's distance from the tunnel exit, where negative values indicate positions inside the tunnel and positive values correspond to positions outside. The tunnel exit is located at 0 m (blue line in Figure 6) and the x-axis denotes individual vehicle indices. It can be observed that vehicles decelerate significantly approximately 180 m before the tunnel exit (black dotted rectangle in Figure 6), accelerate significantly around 80 m before the tunnel exit (red rectangle in Figure 6), and accelerate again about 120 m outside the tunnel (gray rectangle in Figure 6). These observations suggest a common behavioral pattern among drivers when approaching and leaving the tunnel, reflecting the influence of the tunnel exit environment on driving behavior.

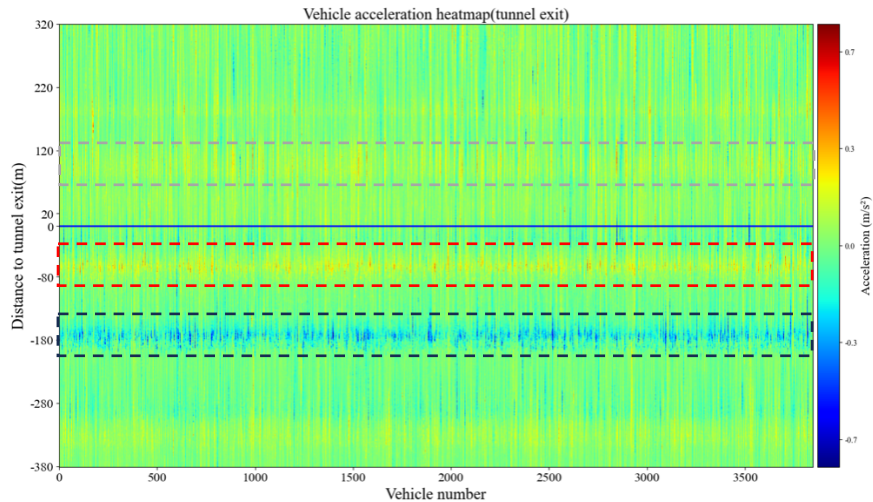

**Figure 6 Acceleration distribution of vehicles at different positions near the tunnel exit**

To quantitatively reflect the environmental influence, this study further extracts acceleration statistics from the tunnel exit section, spanning from 380 m inside to 320 m outside. Specifically, the 20th, 40th, 60th, and 80th percentile values of acceleration are computed at each position along this range, as shown in **Figure 7**. The percentile curves reveal a consistent behavioral pattern among vehicles: significant deceleration begins around -180 m, likely due to the "white hole" effect, where sudden light exposure impairs driver visibility. A slight acceleration follows at approximately -60 m, suggesting drivers begin to adapt to the lighting transition and regain speed. After exiting the tunnel, acceleration increases are observed at roughly 50 and 100 m, after which acceleration stabilizes. The lowest and highest percentile values occur near -150 m and -80 m, respectively.

This spatial pattern of acceleration change provides empirical evidence of the tunnel exit's environmental impact on driving behavior. Accordingly, these percentile-based statistics are integrated into the prediction model as part of the environmental sequence input to capture the influence of environmental changes on vehicle behavior, helping the model to capture location-specific behavioral responses and improve prediction robustness.

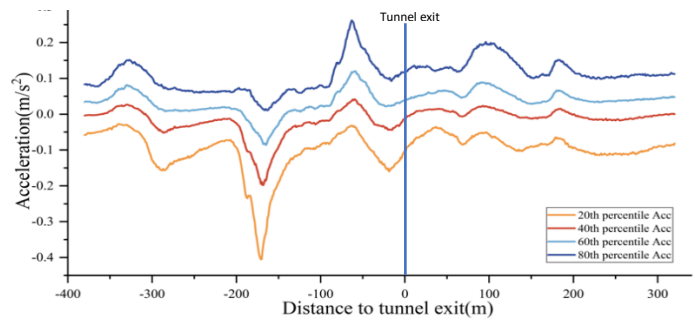

**Figure 7 Vehicle position and acceleration**

To further enhance acceleration prediction accuracy, the relationship between speed and acceleration was examined. A speed-acceleration scatterplot (**Figure 8)** was constructed to investigate how acceleration values are distributed across different speed levels. The plot shows that as speed increases, the range of acceleration initially expands and then contracts, indicating a clear correlation between vehicle speed and acceleration magnitude. Based on this finding, the prediction model incorporates not only acceleration statistics at the prediction point but also the corresponding speed statistics. This additional input provides a more comprehensive representation of vehicle dynamics and improves the model's forecasting capability.

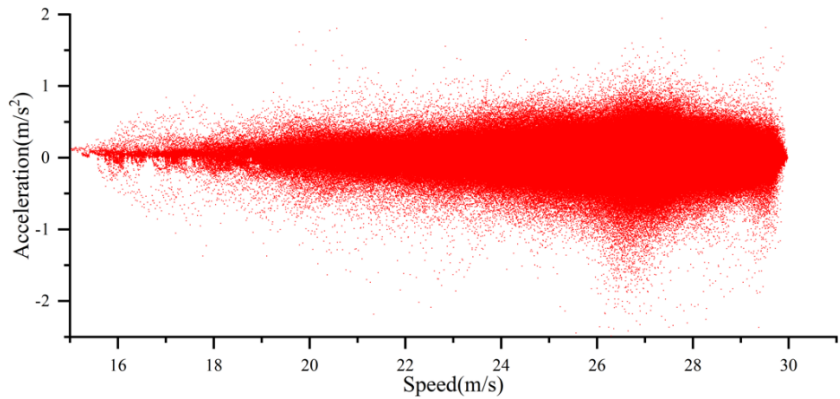

**Figure 8 Scatter plot of speed and acceleration**

## 5.2 Driver classification

Driver acceleration behavior is a key factor of driving style, and the magnitude of acceleration is directly influenced by individual driving tendencies. For example, aggressive drivers typically exhibit greater acceleration and deceleration values, while conservative drivers exhibit lower acceleration variability. Incorporating driver style into the acceleration prediction model can therefore enhance prediction accuracy(Tang et al. 2024).

In this study, trajectories of 3,847 vehicles were collected during the observation period. Three statistical features including acceleration range, average speed, and average acceleration of each vehicle in the observation interval were selected as input variables for clustering. To determine the optimal number of driver categories, the AIC and BIC values were computed for k=1 to k=10, as illustrated in **Figure 9**. Both AIC and BIC reach their minimum when k=3, indicating that a three-cluster solution best fits the data.

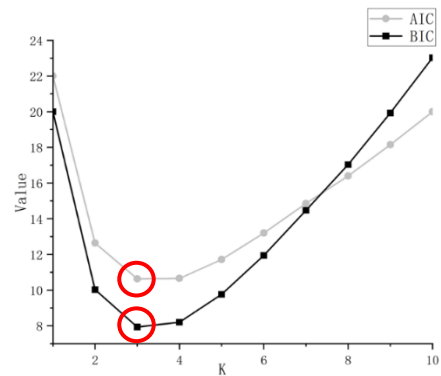

**Figure 9 The relationship between k and AIC and BIC**

**Figure 10** visualizes the distribution of acceleration for three driver categories, and **Table 1** summarizes the average values of their classification features. Based on these results, drivers with lower speeds and limited acceleration fluctuations are classified as conservative; those with moderate speeds and relatively higher sensitivity to acceleration changes are labeled moderate; and those with high speeds and wide acceleration variability are identified as aggressive.

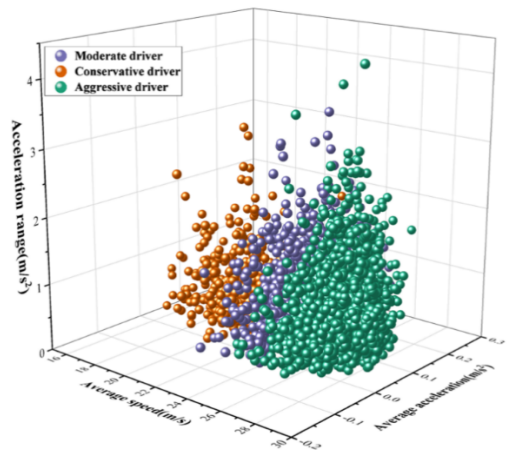

**Figure 10 Driver classification**

Table 1 Statistical characteristics of three types of style data

| Criteria of Classification / Driver Type | Conservative driver | Moderate driver | Aggressive driver |
|---|---|---|---|
| acceleration range($m/s^2$) | 0.8805 | 1.1023 | 1.0383 |
| average acceleration($m/s^2$) | 0.0843 | 0.1198 | 0.1201 |
| average speed($m/s$) | 21.7648 | 25.7631 | 28.0423 |

## 5.3 Data sizes of historical traffic variables

To determine an appropriate input data size for capturing the historical traffic variables at different locations, the proposed acceleration prediction model was used to forecast the acceleration behavior of randomly selected drivers over a 50-meter horizon. The results are presented in Table 2.

As the length of the input historical traffic variables—including speed and acceleration percentiles—increases from 1 minute to 15 minutes, a substantial improvement in prediction accuracy is observed. However, further extending the input duration beyond 15 minutes yields negligible gains. This suggests that a 15-minute window of traffic statistics is sufficient to capture the traffic flow state and represent driver behavior at a given location. Accordingly, this window is used as the environmental sequence input in the prediction model, and the acceleration data from 3,847 vehicles is used to construct the training set.

Table 2 Impact of different sizes of historical traffic variables on prediction accuracy

| Size (min) / Error | 1 | 2 | 15 | 100 | 200 | 300 | 400 |
|---|---|---|---|---|---|---|---|
| MAE | 0.0664 | 0.0620 | 0.0587 | 0.0587 | 0.0591 | 0.0587 | 0.0587 |
| RMSE | 0.0926 | 0.0882 | 0.0821 | 0.0821 | 0.0823 | 0.0818 | 0.0818 |

## 5.4 Acceleration prediction results

To evaluate the effectiveness of the proposed Seq2Seq model for vehicle acceleration prediction, four additional deep learning models were compared, including Recurrent Neural Network(RNN), Artificial Neural Network(ANN), Convolutional Neural Networks(CNN), Bidirectional Long Short-Term Memory(BiLSTM). Three sets of comparative experiments were conducted: (1) prediction experiments using unclustered model and three cluster-specific submodels to assess the impact of driver classification; (2) predictions with and without historical traffic variables to evaluate their contribution to model performance; (3) comparison of prediction accuracy at distances of 10m, 30m, and 50m to examine model robustness across varying prediction horizons.

Acceleration predictions were performed for each of the three driver categories using the Seq2Seq, RNN, ANN, CNN and BiLSTM models. Prediction results over 300 time steps(1 time step represents 0.05s) are shown in **Figure 11**. In these figures, "Y" represents that historical traffic variables was included in the input, while "N" represents it was excluded.

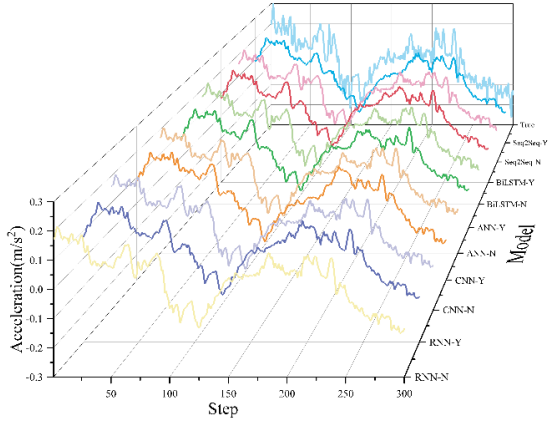 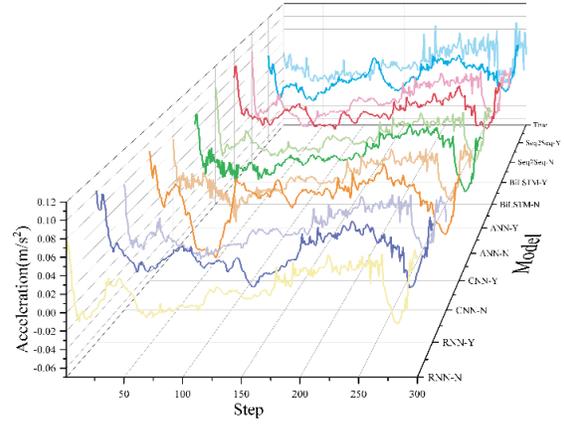

**(a) Prediction for conservative drivers**   **(b) Prediction for moderate drivers**

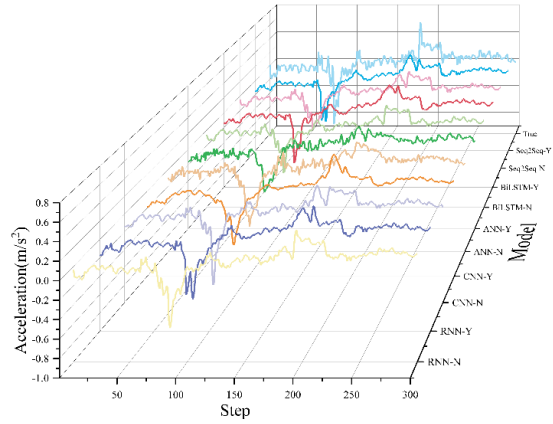

**(c) Prediction for aggressive drivers**
**Figure 11 Prediction for three categories of drivers**

The prediction errors of three categories of drivers with clustered model and unclustered model are shown in **Table 3-4.**

**Table 3   Acceleration prediction error with clustered model**
**(Y:yes, N:No, C:conservative, M:moderate, A:Aggressive)**

| | Prediction distance | 10m | | | | 30m | | | | 50m | | | |
|---|---|---|---|---|---|---|---|---|---|---|---|---|---|
| | Error | MAE | | RMSE | | MAE | | RMSE | | MAE | | RMSE | |
| | Input of historical traffic variables | Y | N | Y | N | Y | N | Y | N | Y | N | Y | N |
| C | Seq2Seq | **0.0383** | **0.0415** | **0.0563** | **0.0625** | **0.0510** | **0.0571** | **0.0733** | **0.0820** | **0.0587** | **0.0667** | **0.0818** | **0.0937** |
| | BiLSTM | 0.0383 | 0.0417 | 0.0563 | 0.0627 | 0.0526 | 0.0579 | 0.0749 | 0.0832 | 0.0597 | 0.0684 | 0.0832 | 0.0956 |
| | RNN | 0.0387 | 0.0418 | 0.0569 | 0.0631 | 0.0522 | 0.0579 | 0.0746 | 0.0830 | 0.0593 | 0.0681 | 0.0828 | 0.0949 |
| | ANN | 0.0406 | 0.0427 | 0.0600 | 0.0641 | 0.0550 | 0.0612 | 0.0781 | 0.0870 | 0.0646 | 0.0723 | 0.0885 | 0.0998 |
| | CNN | 0.0397 | 0.0419 | 0.0583 | 0.0629 | 0.0537 | 0.0587 | 0.0758 | 0.0841 | 0.0604 | 0.0688 | 0.0840 | 0.0960 |
| M | Seq2Seq | **0.0562** | **0.0615** | 0.0862 | **0.0978** | **0.0803** | **0.0887** | **0.1157** | **0.1310** | **0.0911** | **0.1045** | **0.1287** | **0.1485** |
| | BiLSTM | 0.0559 | 0.0616 | **0.0857** | 0.0985 | 0.0807 | 0.0892 | 0.1166 | 0.1315 | 0.0921 | 0.1064 | 0.1309 | 0.1511 |
| | RNN | 0.0566 | 0.0617 | 0.0871 | 0.0987 | 0.0805 | 0.0905 | 0.1167 | 0.1335 | 0.0918 | 0.1066 | 0.1300 | 0.1516 |

|   |       | 0.0594 | 0.0636 | 0.0922 | 0.1019 | 0.0847 | 0.0948 | 0.1219 | 0.1397 | 0.0959 | 0.1118 | 0.1345 | 0.1582 |
|---|-------|--------|--------|--------|--------|--------|--------|--------|--------|--------|--------|--------|--------|
|   | ANN   | 0.0594 | 0.0636 | 0.0922 | 0.1019 | 0.0847 | 0.0948 | 0.1219 | 0.1397 | 0.0959 | 0.1118 | 0.1345 | 0.1582 |
|   | CNN   | 0.0584 | 0.0624 | 0.0897 | 0.0999 | 0.0830 | 0.0916 | 0.1192 | 0.1351 | 0.0929 | 0.1074 | 0.1310 | 0.1528 |
|   | Seq2Seq | **0.0726** | **0.0815** | 0.1234 | **0.1477** | 0.1033 | **0.1199** | 0.1654 | **0.1937** | 0.1214 | **0.1402** | 0.1764 | **0.2164** |
|   | BiLSTM | 0.0728 | 0.0824 | **0.1225** | 0.1488 | 0.1048 | 0.1212 | 0.1641 | 0.1957 | **0.1127** | 0.1419 | **0.1659** | 0.2181 |
| A | RNN   | 0.0737 | 0.0817 | 0.1248 | 0.1486 | **0.1031** | 0.1198 | **0.1615** | 0.1943 | 0.1142 | 0.1412 | 0.1673 | 0.2179 |
|   | ANN   | 0.0804 | 0.0886 | 0.1372 | 0.1580 | 0.1157 | 0.1283 | 0.1819 | 0.2049 | 0.1243 | 0.1472 | 0.1788 | 0.2243 |
|   | CNN   | 0.0756 | 0.0825 | 0.1280 | 0.1501 | 0.1055 | 0.1210 | 0.1666 | 0.1963 | 0.1156 | 0.1414 | 0.1681 | 0.2182 |

Table 4  Acceleration prediction error with unclustered model
(Y:yes, N:No, C:conservative, M:moderate, A:Aggressive)

| Prediction distance | | 10m | | | | 30m | | | | 50m | | | |
|---|---|---|---|---|---|---|---|---|---|---|---|---|---|
| Error | | MAE | | RMSE | | MAE | | RMSE | | MAE | | RMSE | |
| Input of historical traffic variables | | Y | N | Y | N | Y | N | Y | N | Y | N | Y | N |
| C | Seq2Seq | 0.0974 | 0.0984 | 0.1388 | 0.1400 | 0.0982 | 0.0984 | 0.1396 | 0.1401 | 0.0981 | 0.0983 | 0.1396 | 0.1399 |
|   | BiLSTM | 0.0970 | 0.0984 | 0.1384 | 0.1400 | 0.0982 | 0.0984 | 0.1397 | 0.1400 | 0.0981 | 0.0984 | 0.1396 | 0.1399 |
|   | RNN | 0.0972 | 0.0983 | 0.1386 | 0.1400 | 0.0981 | 0.0984 | 0.1396 | 0.1400 | 0.0981 | 0.0984 | 0.1396 | 0.1399 |
|   | ANN | 0.0980 | 0.0983 | 0.1396 | 0.1399 | 0.0981 | 0.0984 | 0.1396 | 0.1400 | 0.0981 | 0.0984 | 0.1396 | 0.1399 |
|   | CNN | 0.0972 | 0.0982 | 0.0784 | 0.1399 | 0.0981 | 0.0980 | 0.1396 | 0.1399 | 0.0980 | 0.0983 | 0.1395 | 0.1398 |
| M | Seq2Seq | 0.1234 | 0.1243 | 0.1763 | 0.1773 | 0.1244 | 0.1246 | 0.1774 | 0.1775 | 0.1245 | 0.1247 | 0.1775 | 0.1778 |
|   | BiLSTM | 0.1233 | 0.1246 | 0.1758 | 0.1778 | 0.1244 | 0.1246 | 0.1774 | 0.1777 | 0.1245 | 0.1247 | 0.1775 | 0.1778 |
|   | RNN | 0.1233 | 0.1246 | 0.1760 | 0.1778 | 0.1244 | 0.1246 | 0.1773 | 0.1777 | 0.1244 | 0.1246 | 0.1774 | 0.1776 |
|   | ANN | 0.1243 | 0.1246 | 0.1773 | 0.1777 | 0.1243 | 0.1246 | 0.1773 | 0.1776 | 0.1244 | 0.1245 | 0.1775 | 0.1775 |
|   | CNN | 0.1238 | 0.1247 | 0.1766 | 0.1779 | 0.1246 | 0.1245 | 0.1776 | 0.1774 | 0.1246 | 0.1242 | 0.1777 | 0.1769 |
| A | Seq2Seq | 0.1209 | 0.1212 | 0.1800 | 0.1804 | 0.1210 | 0.1209 | 0.1800 | 0.1800 | 0.1210 | 0.1212 | 0.1800 | 0.1804 |
|   | BiLSTM | 0.1208 | 0.1212 | 0.1796 | 0.1804 | 0.1210 | 0.1209 | 0.1800 | 0.1800 | 0.1210 | 0.1212 | 0.1800 | 0.1804 |
|   | RNN | 0.1208 | 0.1212 | 0.1796 | 0.1804 | 0.1209 | 0.1209 | 0.1799 | 0.1799 | 0.1210 | 0.1212 | 0.1800 | 0.1804 |
|   | ANN | 0.1211 | 0.1212 | 0.1802 | 0.1804 | 0.1210 | 0.1209 | 0.1800 | 0.1800 | 0.1210 | 0.1212 | 0.1800 | 0.1804 |
|   | CNN | 0.1210 | 0.1211 | 0.1799 | 0.1803 | 0.1210 | 0.1209 | 0.1800 | 0.1799 | 0.1211 | 0.1213 | 0.1800 | 0.1804 |

In Table 3, similar results can be observed that:1) by comparing the prediction errors of models with and without the historical traffic variables, it can be seen that under the same prediction distance and model, the error in the Y column are always smaller than those in the column N, indicating that considering historical traffic variables can improve prediction accuracy. 2)by comparing different prediction distances, with the same prediction model and historical traffic variables, it can be observed that as the prediction distance increases, the error also increases. The MAE and RMAE both show similar trends, with the largest error occurring at a distance of 50 meters.

The performance of different prediction models for data from drivers with different driving styles was compared. In **Table** 3, for conservative drivers, the Seq2Seq consistently demonstrates the optimal performance among five prediction mdoels. For moderate drivers, the Seq2Seq model exihibits the best performance in nearly all scenarios except for the column 10m-RMSE-Y. Even in this situation, the Seq2Seq achieves the second-best performance. For aggressive drivers, the optimal model is not definitive, but the Seq2Seq model shows the best performace in most scenarios. As shown in Table 3, the Seq2Seq model demonstrates the best overall performance across all driver categories, with particularly strong results for conservative drivers. Figure 12 further visualizes this trend, confirming that Seq2Seq consistently outperforms the baseline models in most prediction scenarios.

The impact of incorporating historical traffic variables—capturing the behavior of multiple vehicles at specific spatial locations—was also examined. Results show that including such information consistently enhances prediction accuracy across all scenarios. In general, the benefits of incorporating historical traffic variables are most pronounced for aggressive drivers, moderate for moderate drivers, and least significant for conservative drivers. This trend aligns with the broader acceleration distribution ranges observed across these driver categories. These findings highlight that leveraging aggregated behavioral patterns from multiple vehicles at the target location can significantly improve model robustness, especially in the presence of unpredictable driving behavior.

The impact of driver clustering was evaluated by comparing model performance with and without classification. Comparing Tables 3 and Table 4, clustering consistently enhances prediction accuracy across all models. On average, clustering improved prediction accuracy by 20% for conservative drivers, 20% for moderate drivers, and 23% for aggressive drivers. These results confirm that accounting for individual driving styles through clustering enables models to better capture behavioral heterogeneity, leading to more accurate predictions.

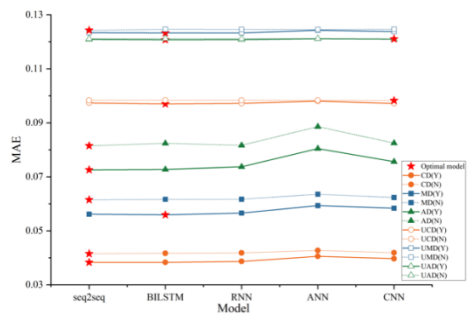

(a) MAE of 10-meter prediction

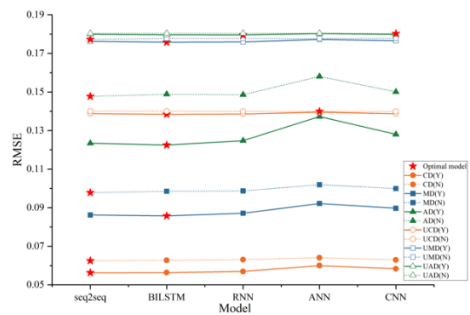

(b) RMSE of 10-meter prediction

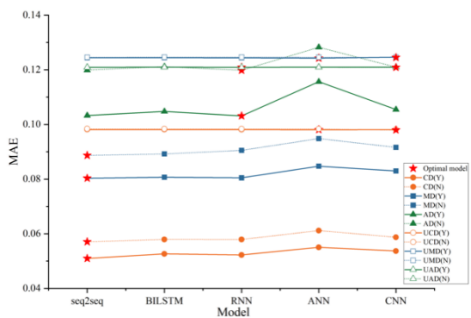

(c) MAE of 30-meter prediction

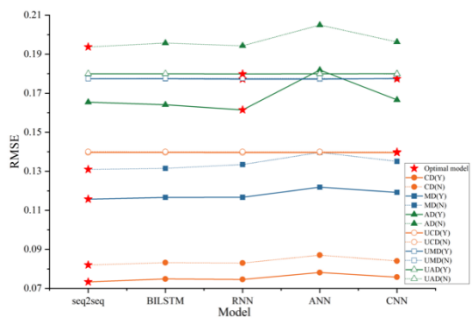

(d) RMSE of 30-meter prediction

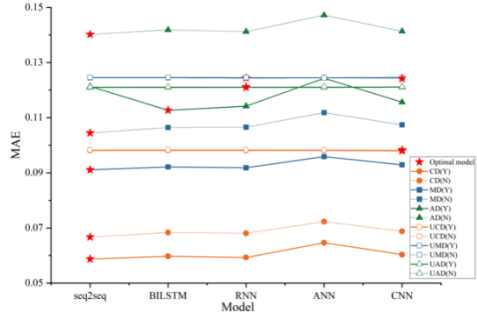
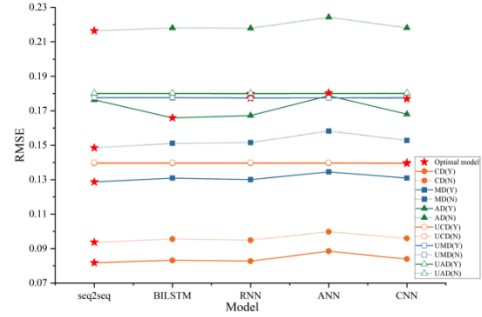

(e) MAE of 50-meter prediction    (f) RMSE of 50-meter prediction

**Figure 12** Prediction error of five models(the red star represents the model with the best performance among the five models; the orange circle, blue square and green triangle represent the Conservative driver (CD), Moderate driver (MD) and Aggressive driver (AD), respectively. Solid line (Y) and dotted lines (N) represent the presence or absence of historical traffic variables; Solid markers indicate driver driver classification, while hollow markers represent unclassified (U) drivers.)

## 6. Discussion

(1) Vehicle behavior prediction. Most vehicle behavior prediction studies emphasize speed prediction, while acceleration prediction remains underexplored—only (Zou et al. 2022) has touched on this aspect. However, acceleration prediction poses greater challenges: it requires higher sampling resolution, greater model responsiveness to subtle variations, and more accurate input data. While speed prediction supports macroscopic traffic planning, acceleration prediction is essential for microscopic tasks such as adaptive cruise control, emergency braking, and energy-efficient vehicle operations. In intelligent vehicle applications, particularly connected and autonomous vehicles (CAVs), anticipating the acceleration of surrounding human-driven vehicles allows for earlier response, improved ride comfort, and reduced energy consumption—benefits that are especially valuable in complex road environments like tunnels and curves.

(2) Prediction models. Deep learning has become a mainstream approach for traffic behavior modeling, with LSTM (Zou et al. 2022; Yeon et al. 2019; Meng et al. 2022; Geng et al. 2023; Li, Cheng, and Ge 2023) and GRU (Zou et al. 2022; Li, Cheng, and Ge 2023) being the most widely used. In contrast to these recurrent models, this study introduces a Seq2Seq framework enhanced with attention mechanisms, specifically designed for multi-step acceleration prediction. To our knowledge, this is the first study to apply an attention-enhanced Seq2Seq model to vehicle acceleration prediction. Experimental comparisons confirm that Seq2Seq consistently outperforms BiLSTM, RNN, ANN, and CNN, highlighting its capacity to model complex temporal dynamics and local variations in driver behavior.

(3) Tunnel vehicle behavior prediction. Previous studies on tunnel scenarios have predominantly focused on regional average speed prediction. For example, (Xu et al. 2022)proposed an expected speed model for vehicle groups based on environmental transitions at spiral tunnel exits. Similarly, (Cao, Li, and Chan 2020)and(Xiao, Liang, and Chen 2022)estimated average speeds within tunnel segments or at specific zones such as entrances and emergency areas. However, these studies primarily focus on macroscopic group-level speed estimation and overlook the fine-grained behavior of individual vehicles, especially their continuous acceleration patterns, which are critical for vehicle-level control and safety applications. In contrast, this study captures the microscopic acceleration response of individual vehicles to tunnel exit environments, providing more detailed behavioral insights. These vehicle-level findings may serve as a valuable supplement to existing macroscopic models by informing the development of more precise speed adjustment factors in the Highway Capacity Manual (HCM)—especially for tunnel exits where abrupt behavioral transitions occur due to lighting and geometric changes.

(4) Clustering effects. This study also investigated how incorporating driver classification affects prediction performance. While ([Li, Cheng, and Ge 2023](#)) found that conservative drivers achieve the highest accuracy, followed by aggressive and moderate drivers, our results indicate a slightly different pattern: conservative drivers still perform best, but moderate drivers outperform aggressive ones. This discrepancy may arise from differences in behavioral variability and data distribution. As shown in Figure 10, moderate drivers have an average speed of 25.76 m/s, while aggressive drivers average 28.04 m/s. As previously shown in Figure 8, acceleration variability increases with speed up to 26–27 m/s before narrowing. This suggests that vehicles clustered near the speed limit exhibit greater behavioral fluctuations. In contrast, drivers exceeding the speed limit may maintain steadier speeds and smoother driving patterns. These findings reveal that prediction difficulty is not linearly correlated with driving aggressiveness, but is influenced by both speed-level concentration and behavioral volatility.

## 7. Conclusions

This study proposes a vehicle acceleration prediction framework that jointly incorporates environmental sequences and individual driving behavior sequences. By leveraging high-frequency radar and video data collected from the Guangzhou Baishi Tunnel, the model captures both the historical traffic variables at specific spatial locations and the individual vehicle behavior. A Seq2Seq network enhanced by attention mechanisms is developed to perform multi-step acceleration forecasting under dynamic traffic conditions.

A key innovation of this work is the integration of both environmental influences reflected by historical traffic variables and individual vehicle driving behavior sequence into a unified deep learning model. This dual-input approach allows the model to account for general driving trends shaped by the environment as well as individual driving behavior dynamics, significantly improving robustness and accuracy—especially over longer prediction horizons.

Based on driving behavior indicators—including average speed, acceleration range, and acceleration mean—drivers are clustered into three categories: conservative, moderate, and aggressive. Conservative drivers exhibit low speeds and stable acceleration patterns; moderate drivers show greater variability in response; and aggressive drivers are characterized by high speeds and frequent acceleration changes. Prediction performance varies significantly across groups, with conservative drivers achieving the highest accuracy due to the lower variability and higher regularity of their driving behavior.

Experimental results show that the proposed Seq2Seq model consistently outperforms conventional deep learning baselines (RNN, ANN, CNN, BiLSTM), particularly when incorporating traffic history variables. Moreover, the inclusion of driver classification enhances model adaptability to behavioral diversity, further boosting prediction performance.

In addition to the modeling contributions, this study provides practical insights for traffic engineering. Specifically, the results may serve as a valuable reference for refining speed adjustment factors in the Highway Capacity Manual (HCM), especially for tunnel exits where abrupt behavioral changes occur due to environmental transitions.

Looking ahead, two directions will be prioritized in future work. First, to enhance the model's generalization, we will extend the proposed framework to additional traffic scenarios such as intersections, tunnel entrances, and highway ramps. Second, improved algorithms and model frameworks will be designed to further enhance the accuracy of long-range acceleration prediction. These advancements will support the broader application of high-fidelity behavior prediction in intelligent traffic management and autonomous vehicle control.

# Acknowledgments

This study was supported by the Shaanxi Province Postdoctoral Funding Project (grant number 2023BSHEDZZ216), Qinchuangyuan Cites High-level Innovation and Entrepreneurship Talent Project (grant number QCYRCXM2023-110), General funding project of China Postdoctoral Science Foundation (2024M752739), Key R&D Program Project of Shaanxi Province(2024GX-ZDCYL-02-14), Natural Science Foundation Project (grant number 52362050).

# Author contributions

The authors confirm contribution to the paper as follows: study conception and design: Lexing Zhang, Wenxuan Wang; data collection: Wenxuan Wang; code writing: Lexing Zhang, Jiale Lei; analysis and interpretation of results: Lexing Zhang, Jiale Lei; draft manuscript preparation: Jiale Lei, Lexing Zhang, Wenxuan Wang, Yin Feng, Hengxu Hu. All authors reviewed the results and approved the final version of the manuscript. Wenxuan Wang and Lexing Zhang contributed equally to this paper.

# References

Bengio, Samy, Oriol Vinyals, Navdeep Jaitly, and Noam Shazeer. 2015. "Scheduled Sampling for Sequence Prediction with Recurrent Neural Networks." In *29th Annual Conference on Neural Information Processing Systems (NIPS)*. Montreal, CANADA.

Cao, Miaomiao, Victor O. K. Li, and Vincent W. S. Chan. 2020. "A CNN-LSTM Model for Traffic Speed Prediction." In *2020 IEEE 91st Vehicular Technology Conference (VTC2020-Spring)*, 1-5. Antwerp, Belgium.

Geng, Qingtian, Zhi Liu, Baozhu Li, Chen Zhao, and Zhijun Deng. 2023. 'Long-Short Term Memory-Based Heuristic Adaptive Time-Span Strategy for Vehicle Speed Prediction', *IEEE Access*, 11: 65559-68.

Jiang, B. N., and Y. S. Fei. 2014. "On-road PHEV Power Management with Hierarchical Strategies in Vehicular Networks." In *2014 IEEE Intelligent Vehicles Symposium Proceedings*, 1077-84. Dearborn, MI.

Jiang, Bingnan, and Yunsi Fei. 2017. 'Vehicle Speed Prediction by Two-Level Data Driven Models in Vehicular Networks', *IEEE Transactions on Intelligent Transportation Systems*, 18: 1793-801.

Jones, Ian, and Kyungtae Han. 2019. "Probabilistic Modeling of Vehicle Acceleration and State Propagation with Long Short-Term Memory Neural Networks." In *30th IEEE Intelligent Vehicles Symposium (IV)*, 2236-42. Paris, FRANCE.

Krajewski, Robert and Bock, Julian and Kloeker, Laurent and Eckstein, Lutz. 2018. "The highD Dataset: A Drone Dataset of Naturalistic Vehicle Trajectories on German Highways for Validation of Highly Automated Driving Systems." In *2018 21st International Conference on Intelligent Transportation Systems (ITSC)*, 2118-25.

Li, Qinyin, Rongjun Cheng, and Hongxia Ge. 2023. 'Short-Term Vehicle Speed Prediction based on BiLSTM-GRU Model Considering Driver Heterogeneity', *Physica a-Statistical Mechanics and Its Applications*, 610.

Li, Yaguang, Rose Yu, Cyrus Shahabi, and Yan Liu. 2017. 'Diffusion Convolutional Recurrent Neural Network: Data-Driven Traffic Forecasting'.

Li, YuFang, MingNuo Chen, XiaoDing Lu, and WanZhong Zhao. 2018. 'Research on Optimized GA-SVM Vehicle Speed Prediction Model Based on Driver-Vehicle-Road-Traffic System', *Science China Technological Sciences*, 61: 782-90.

Luo, Haoxuan, Xiao Hu, and Linyu Huang. 2023. 'A Hybrid Model for Vehicle Acceleration Prediction', *Sensors*, 23.

Luong, Minh-Thang. 2015. 'Effective Approaches to Attention-based Neural Machine Translation'.

Lv, Yisheng, Yanjie Duan, Wenwen Kang, Zhengxi Li, and Fei-Yue Wang. 2014. 'Traffic Flow Prediction with Big Data: A Deep learning Approach', *IEEE Transactions on Intelligent Transportation Systems*, 16: 865-73.

Ma, Xiaolei, Zhimin Tao, Yinhai Wang, Haiyang Yu, and Yunpeng Wang. 2015. 'Long Short-Term Memory Neural Network for Traffic Speed Prediction Using Remote Microwave Sensor Data', *Transportation research part C: emerging technologies*, 54: 187-97.

Meng, Xianwei, Hao Fu, Liqun Peng, Guiquan Liu, Yang Yu, Zhong Wang, and Enhong Chen. 2022. 'D-LSTM: Short-Term Road Traffic Speed Prediction Model Based on GPS Positioning Data', *IEEE Transactions on Intelligent Transportation Systems*, 23: 2021-30.

"NGSIM [WWW Document]." In. 2006.

Raza, N., M. A. Habib, M. Ahmad, Q. Abbas, M. B. Aldajani, and M. A. Latif. 2024. 'Efficient and Cost-Effective Vehicle Detection in Foggy Weather for Edge/Fog-Enabled Traffic Surveillance and Collision Avoidance Systems', *Cmc-Computers Materials & Continua*, 81: 911-31.

Shi, Qi, and Mohamed Abdel-Aty. 2015. 'Big Data applications in real-time traffic operation and safety monitoring and improvement on urban expressways', *Transportation Research Part C-Emerging Technologies*, 58, Part B: 380-94.

Su, Jianyu, Peter A Beling, Rui Guo, and Kyungtae Han. 2020. "Graph Convolution Networks for Probabilistic Modeling of Driving Acceleration." In *23rd IEEE International Conference on Intelligent Transportation Systems (ITSC)*, 1-8. ELECTR NETWORK.

Sun, Shiliang, Changshui Zhang, and Yi Zhang. 2017. 'Traffic Flow Forecasting Using a Spatio-Temporal Bayesian Network Predictor'.

Tang, Shuning, Yajie Zou, Hao Zhang, Yue Zhang, and Xiaoqiang Kong. 2024. 'Application of CNN‐LSTM Model for Vehicle Acceleration Prediction Using Car‐following Behavior Data', *Journal of Advanced Transportation*, 2024.

Vaswani, A. 2017. "Attention is all you need." In *31st Annual Conference on Neural Information Processing Systems (NIPS)*. Long Beach, CA.

Wang, Wenxuan, Bo Yan, Xiaojun Li, Minghao Tian, Rong Jiao, and Hua Zhang. 2024. 'Multiple Pedestrian Tracking Using LiDAR Network in Complex Indoor Scenarios', *IEEE Sensors Journal*: 13175-92.

Wei, Hua, Guanjie Zheng, Huaxiu Yao, and Zhenhui Li. 2018. "Intellilight: A Reinforcement Learning Approach for Intelligent Traffic Light Control." In *24th ACM SIGKDD Conference on Knowledge Discovery and Data Mining (KDD)*, 2496-505. London, ENGLAND.

Williams, Billy M, and Lester A Hoel. 2003. 'Modeling and Forecasting Vehicular Traffic Flow as a Seasonal ARIMA Process: Theoretical Basis and Empirical Results', *Journal of transportation engineering*, 129: 664-72.

Xiao, Yao, Bo Liang, and Kai Chen. 2022. 'Analysis and Prediction Models for Operating Speed of Vehicles in Expressway Superlong Tunnels based on Geometric and Traffic rRelated Parameters', *Traffic Injury Prevention*, 23: 410-15.

Xu, Xiaoling, Xuejian Kang, Xiaoping Wang, Shuai Zhao, and Chundi Si. 2022. 'Research on Spiral Tunnel Exit Speed Prediction Model Based on Driver Characteristics', *Sustainability*, 14.

Yan, B., X. Y. Zhao, N. Xu, Y. Chen, and W. B. Zhao. 2019. 'A Grey Wolf Optimization-Based Track-Before-Detect Method for Maneuvering Extended Target Detection and Tracking', *Sensors*, 19.

Yan, Bo, Andrea Giorgetti, and Enrico Paolini. 2021. 'A Track-Before-Detect Algorithm for UWB Radar Sensor Networks', *Signal Processing*, 189.

Yan, Bo, Enrico Paolini, Luping Xu, and Hongmin Lu. 2022. 'A Target Detection and Tracking Method for Multiple Radar Systems', *IEEE Transactions on Geoscience and Remote Sensing*, 60: 1-21.

Yan, Bo, Enrico Paolini, Na Xu, Zhifeng Sun, and Luping Xu. 2021. 'Multiple Maneuvering Extended Targets Detection by 3D Projection and Tracklet Association', *Signal Processing*, 179.

Yeon, Kyuhwan, Kyunghan Min, Jaewook Shin, Myoungho Sunwoo, and Manbae Han. 2019. 'EGO-Vehicle Speed Prediction Using a Long Short-Term Memory based Recurrent Neural Network', *International Journal of Automotive Technology*, 20: 713-22.

Yu, Zhou, Xingyu Shi, and Zhaoning Zhang. 2023. 'A Multi-Head Self-Attention Transformer-based model for Traffic Situation Prediction in Terminal Areas', *IEEE Access*, 11: 16156-65.

Yuan, Tian, Xuan Zhao, Rui Liu, Qiang Yu, Xichan Zhu, Shu Wang, and Karl Meinke. 2023. 'Driving Intention Recognition and Speed Prediction at Complex Urban Intersections Considering Traffic Environment', *IEEE Transactions on Intelligent Transportation Systems*: 4470-88.

Zou, Yajie, Lusa Ding, Hao Zhang, Ting Zhu, and Lingtao Wu. 2022. 'Vehicle Acceleration Prediction Based on Machine Learning Models and Driving Behavior Analysis', *Applied Sciences-Basel*, 12.